\documentclass[11pt,a4paper]{article}
\pdfoutput=1
\usepackage[hyperref]{emnlp-ijcnlp-2019}
\usepackage{times}
\usepackage{latexsym}
\usepackage{graphicx}
\usepackage{amsfonts}
\usepackage{amsmath}
\usepackage{multirow}
\usepackage{epstopdf}
\usepackage{CJK}
\usepackage{url}

\title{Investigating Self-Attention Network for Chinese Word Segmentation}

\author{Leilei Gan \\
  Westlake University \\
  {ganleilei@westlake.edu.cn} \\\And
  Yue Zhang \\
  Westlake University \\
  {zhangyue@westlake.edu.cn} \\}

\date{}

\begin{document}
\begin{CJK*}{UTF8}{gkai}
\maketitle
\begin{abstract}
Neural network has become the dominant method for Chinese word segmentation. Most existing models cast the task as sequence labeling, using BiLSTM-CRF for representing the input and making output predictions. Recently, attention-based sequence models have emerged as a highly competitive alternative to LSTMs, which allow better running speed by parallelization of computation. We investigate self attention for Chinese word segmentation, making comparisons between BiLSTM-CRF models. In addition, the influence of contextualized character embeddings is investigated using BERT, and a method is proposed for integrating word information into SAN segmentation. Results show that SAN gives highly competitive results compared with BiLSTMs, with BERT and word information further improving segmentation for in-domain and cross-domain segmentation. Our final models give the best results for 6 heterogenous domain benchmarks.
\end{abstract}

\section{Introduction}
% 中文分词的重要性，以及许多基于RNN/CNN的方法。
Word segmentation is a necessary pre-processing step for Chinese language processing \cite{zhang2007chinese, sun2011enhancing, jiang2013discriminative, xu2016dependency, cai2017fast}. The dominant method treats Chinese Word Segmentation (CWS) as a sequence labeling problem \cite{xue2003chinese}, where neural network models \cite{yang2017neural,zhou2017word,zhang2016transition,cai2016neural} have achieved the state-of-the-art results. A representative model \cite{chen2015long,chen2015gated} takes LSTM \cite{hochreiter1997long} as a feature extractor, and a standard CRF \cite{lafferty2001conditional} layer is used on top of a BiLSTM layer to predict the label sequences.% input character representation

Recently, self-attention network (SAN) \cite{vaswani2017attention} has been shown effectiveness for a range of natural language processing tasks, such as machine translation \cite{tang2018self}, constituency parsing \cite{kitaev2018constituency}, and semantic role labeling \cite{tan2018deep}. Compared with recurrent neural networks (RNNs) \cite{elman1990finding}, SAN has advantages of capturing long-term dependencies and supporting parallel computing more easily. However, its effectiveness on CWS has not been fully investigated in the literature.

%to the best of our knowledge, no related research has been done for neural CWS. In this paper, we investigate the influence of global and local self-attention on Chinese word segmentation and propose a local self-attention based model for CWS\footnote{Code and trained models will be available soon}. 

%We also investigate the effectiveness of contextualized character representation both in in-domain and cross-domain CWS tasks. Further more, we propose a neural type-supervised domain adaptation method which can help to handle out of vocabulary words in cross-domain evaluation by generalizing words to their learned pos tag emebedding.

%However, there are two defects in this model: i) the inherent sequential computation can hardly be paralleled. ii) it can be difficult for LSTM to capture long-term dependencies. 

%trained from large-scale corpus have demonstrated huge advantages over traditional context-independent word representation \cite{mikolov2013distributed,pennington2014glove} in representing the meaning of words

We empirically investigate SAN for CWS by building a SAN-CRF word segmentor, studying the influence of global and local attention for segmentation accuracy. Based on the SAN-CRF segmentation model, we investigate two further questions. First, in Chinese, characters are highly polysemantic and the same character can have different meanings in different context. SAN has also been shown a useful method for training contextualized word representations \cite{devlin2018bert,radford2018improving}. We compare context-independent character representations \cite{mikolov2013distributed,pennington2014glove} with contextualized character representations in both in-domain and cross-domain CWS evaluation. 

Second, out of vocabulary words, especially domain specific noun entities, raises a challenge for cross-domain CWS. To solve this problem, domain lexicons can be used \cite{zhang2014type, zhang2018neural} for cross-domain CWS tasks. We consider a novel method for integrating lexicons to SAN for cross-domain CWS, using attention to integrate word information by generalizing words into POS tags, resulting in end-to-end neural type-supervised domain adaptation.

%Moreover, SAN also allows us to easily integrate range of information collected from large-scale texts in two ways. First, it has been proven that word information can be beneficial for CWS \cite{zhang2016transition, cai2016neural}. In this paper, different from previous work, for each character in sentence, we incorporate word information by attending to all matching words from lexicon. Therefore, each character representation is enhanced with the weighted sum of all candidate word embedding, while weights are the probabilities calculated by attending character to all candidate words.

%firstly, we use attention mechanism to incorporate matched word information from external lexicon to enhance the representation of character. Secondly, there still may be domain specific words which may not be covered by external lexicon. Therefore, 

%For example, as shown in Figure \ref{Model Architecture}, 
%\begin{CJK*}{UTF8}{nsung}
%	the character "中"(Center) matches two words from lexicon, "中国"(China) and "中国科学"(Chinese Science). Intuitively, the importance of two words for character "中"(Center) is different, and the probability of "中"(Center) segmented to "中国"(China) is higher than segmented to "中国科学"(Chinese Science). 
%\end{CJK*}
%

% Contributions
Results on three benchmarks show that SAN-CRF can achieve competitive performance compared with BiLSTM-CRF. In addition, BERT character embeddings are used for both in-domain and cross-domain evaluation. In cross-domain evaluation, the proposed neural type-supervised method gives an averaged error reduction of 30.32\% on three cross-domain datasets. Our method gives the best results on standard benchmarks including CTB, PKU, MSR, ZX, FR and DL. To the best of our knowledge, we are the first to investigate SAN for CWS\footnote{Code and trained models will be made available}.
%the contributions of our paper are as follows:
%\begin{itemize}
%	\item To the best of our knowledge, we are the first to propose deep Transformer based CWS method.
%	\item We use contextualized character representation and archive the state-of-art performance in many in-domain datasets.
%	\item We use attention mechanism to incorporate additional dictionary knowledge and archive the state-of-art performance in many cross-domain datasets.
%\end{itemize}
\label{sec:model}
\begin{figure}[t]
	\begin{flushleft}
	\includegraphics[width=0.48\textwidth]{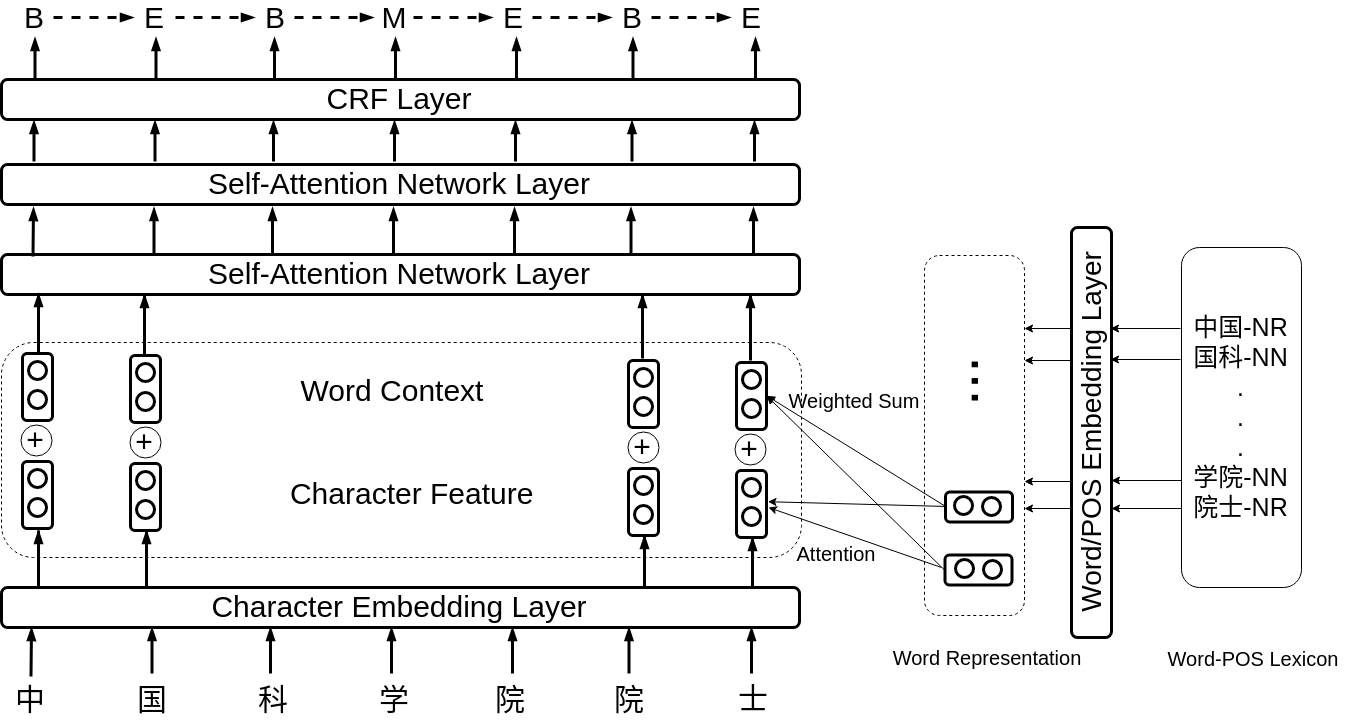}
	\caption{Model Overview}
	\label{Model Architecture}
	\end{flushleft}
\end{figure}

\section{Baseline}
We take BiLSTM-CRF as our baseline, which has been shown giving the state-of-the-art results \cite{chen2015long, yang2018subword}. Formally, given an input sentence with $m$ characters $s=c_1, c_2, ..., c_m$, where $c_i$ denotes the $i$th character, the task of character based CWS is to assign each character $c_i$ with a label $y_i$, where $y_i \in \{B, M, E, S\}$ \cite{xue2003chinese}. The label $B, M, E$ and $S$ represent the begin, middle, end of a word and single character word, respectively.

For each character $c_i$, its input representation $\textbf{e}_{i}$ is the concatenation of unigram character embedding $\textbf{e}_{c_i}$ and bigram character embedding $\textbf{e}_{c_ic_{i+1}}$. Following \cite{chen2015long}, we take BiLSTM to capture the context of character in both forward and backward directions.

The forward hidden state $\overrightarrow{\mathbf{h}}_i$ of character $c_i$ is calculated as follows:
\begin{equation}
\begin{aligned}
\mathbf{i}_i \ &= \ \sigma(\textbf{W}_i\textbf{h}_{i-1} + \textbf{U}_i\textbf{e}_{i} + \textbf{b}_i) \\
\mathbf{f}_i \ &= \ \sigma(\textbf{W}_f\textbf{h}_{i-1} + \textbf{U}_f\textbf{e}_{i} + \textbf{b}_f) \\
\mathbf{o}_i \ &= \ \sigma(\textbf{W}_o\textbf{h}_{i-1} + \textbf{U}_o\textbf{e}_{i} + \textbf{b}_o) \\
\mathbf{\widetilde{c}}_{i} \ &= \ \tanh(\textbf{W}_{\widetilde{c}}\textbf{h}_{i-1} + \textbf{U}_{\widetilde{c}}\textbf{e}_{i} + \textbf{b}_{\widetilde{c}}) \\
\mathbf{c}_i \ &= \ \mathbf{f}_i \odot \mathbf{c}_{i - 1} + \mathbf{i}_i \odot \mathbf{\widetilde{c}}_{i} \\
\overrightarrow{\mathbf{h}}_i \ &= \ \mathbf{o}_i \odot \tanh(\mathbf{c}_{i})
\end{aligned}
\end{equation}
where $\mathbf{i}_i$, $\mathbf{f}_i$ and $\mathbf{o}_i$ are input, forget and output gates, respectively. $\sigma$ and $\odot$ are element-wise sigmoid and product function respectively while $\textbf{W}$, $\textbf{U}$ and $\textbf{b}$ are model parameters to learn.

The backward hidden state $\overleftarrow{\mathbf{h}}_i$ can be obtained in a similar way. Thus, the hidden state of character $c_i$ is the concatenation of $\overleftarrow{\mathbf{h}}_i$ and $\overrightarrow{\mathbf{h}}_i$:
\begin{equation}
\mathbf{h}_i = \overleftarrow{\mathbf{h}}_i \oplus \overrightarrow{\mathbf{h}}_i
\end{equation}

In the scoring layer, a CRF is used to consider the dependencies of adjacent labels. The probability of a label sequence $y=y_1, y_2, ..., y_n$ of sentence $s$ is:
\begin{equation}
P(y|s) = \dfrac{\textrm{exp}(\sum\limits_{i=1}^{n}(F(y_i) + L(y_{i-1}, y_i)))}{\sum\limits_{y^{'} \in \mathbb{C}(s)}\textrm{exp}(\sum\limits_{i=1}^{n}(F(y^{'}_i) + L(y^{'}_{i-1}, y^{'}_i)))}
\end{equation}
where $\mathbb{C}(s)$ is the set of all possible label sequences of sentence $s$. $F(y_i)$ is the emission score of $y_i$ and $L(y_{i-1}, y_{i})$ is the transition score from $y_{i - 1}$ to $y_{i}$. 
\section{Model}
Figure \ref{Model Architecture} shows our segmentor framework on a input character sequence ``(Fellow of the Chinese Academy of Sciences)". The model takes character representation and positional embeddings as input. By matching the input to a word-POS lexicon, word information is investigated by using attention for each character. Multiple layers of self-attention network \cite{vaswani2017attention} are used as feature extractor to replace BiLSTM in the baseline. Similar to the baseline, we also use a CRF layer on top of the self-attention network to model the dependencies of adjacent labels.
\subsection{Embedding Layer}
As shown in Figure \ref{Model Architecture}, the embedding layer consists of character embeddings and positional embeddings. The character representation of $c_i$ is the concatenation of unigram character embedding $\textbf{e}_{c_i}$ and bigram character embedding $\textbf{e}_{c_ic_{i+1}}$.
\begin{equation}
\widetilde{\mathbf{x}}_i^c = \textbf{e}_{c_i} \oplus \textbf{e}_{c_ic_{i+1}}
\end{equation}
where $\oplus$ represents concatenation operation.
% positional embedding

Because self-attention network does not explicitly consider sequence information, positional embedding is added to the input of self-attention network as follows:
\begin{equation}
\begin{aligned}
PE_{(pos, 2i)} &=  sin(pos/10000^{2i/d_{model}}) \\ 
PE_{(pos, 2i+1)} &= sin(pos/10000^{2i/d_{model}}) \\
\mathbf{x}_i^c &= \widetilde{\mathbf{x}}_i^c + PE
\end{aligned}
\end{equation}
where \textit{pos} is the position, $i$ is the dimension, $d_{model}$ is the dimension of output, respectively, and $+$ denotes vector addition.
\subsection{Self-Attention Network}
We extend the model of \citet{vaswani2017attention} for the SAN segmentor. The model has multiple identical layers, each of which is composed of a multi-head self-attention sub-layer and a position-wise fully connected feed-forward network.
% multi-head attention

Multi-head self-attention is used to exchange information directly between positions in the sequence. First, for single-head self-attention, the representations of a sequence $X$ is computed by scaled dot-product attention as follows:
\begin{equation}
\begin{aligned}
\textrm{Attention}(X) & = \textrm{Attention}(Q, K, V) \\
                      & =\textrm{softmax}(\dfrac{QK^T}{\sqrt{d_k}})V
\end{aligned}
\end{equation}
\label{single-head self-attention}where $Q=W^T_QX, K=W^T_KX, V=W^T_VX$ are query, key and value vectors, respectively, and $W^T_Q,\ W^T_K,\ W^T_V$ are parameters.

\noindent \textbf{Local Self-Attention} \quad In order to investigate the effect of long-term dependencies on CWS task, we propose a \textit{local self-attention}, which only attends to surrounding positions for each character instead of all positions in the sequence. The intuition is that long-term dependencies may bring more noise than information in a sequence labeling task \cite{luong2015effective}. The \textit{local self-attention} is denoted as:
\begin{equation}
\begin{aligned}
\textrm{L-Attention}(X) & = \textrm{Attention}(Q, K, V) \\
& =(\textrm{softmax}(\dfrac{QK^T}{\sqrt{d_k}}) \odot W) V \\
\end{aligned}
\end{equation}
where $W$ is a matrix to control the self-attention inner a window and its element $W_{ij}$ is denoted as:
\begin{equation}
	W_{ij} =
	\begin{cases} 
	1& j - i <= WS \\ 
	-\infty& \text{otherwise}\\
	\end{cases}
\end{equation}
Here $WS$ is the window size. 

\noindent \textbf{Multi-Head Self-Attention} is used, which linearly maps $Q$, $K$ and $V$ into multiple versions $Q_i$, $K_i$ and $V_i$ and then concatenates the outputs of different $\textrm{head}_i$ as follows:
\begin{equation}
\textrm{MH}(X) = \textrm{Concat}(\textrm{head}_1, ..., \textrm{head}_n)W^o \\
\end{equation}
where 
\begin{equation}
\textrm{head}_i = \textrm{Attention}(QW_i^Q, KW_i^K, VW_i^V)
\end{equation} 
$W^o$, $W_i^Q$, $W_i^K$ and $W_i^V$ are parameters.
% feed forward network

On top of the multi-head attention sub-layer, a fully connected feed-forward network (FNN) is applied to each position. FNN is composed of two linear transformations with a ReLU activation.
\begin{equation}
\textrm{FNN}(x) = W_2\textrm{ReLU}(0, xW_1 + b_1) + b_2
\end{equation}

\section{Rich Character and Word Features}
We incorporate rich character and word features into SAN model. Specifically, pre-trained contextualized character representation is introduced as well as a word-based neural type-supervised domain adaptation method.

\subsection{BERT Character Representation}
BERT \cite{devlin2018bert} is trained from a large scale corpora by a deep bidirectional Transformer using masked LM tasks. Usages of BERT can be divided into feature-based and fine-tuning methods. The former fixes all model parameters and directly extracts character features from the pre-trained model, while the latter jointly fine-tunes all parameters on downstream tasks. We take the latter method, feeding the input sequence of characters into BERT and use the top layer output as character representation. Development experiments show that fine-tuning BERT embeddings give higher results than the feature-based method.

Formally, character $c_i$ is represented using pre-trained BERT embedding according to the whole sentence.
\begin{equation}
\mathbf{e}_i^c = \mathbf{e}^c(c_i)
\end{equation}
where $\mathbf{e}^c$ denotes a pre-trained BERT character embedding.

\subsection{Integrating Word-POS Lexicon for Type-Supervision}
We integrate word information into SAN to handle rare words in cross-domain settings. Following the definition by \citet{zhang2014type}, we describe this model in a cross-domain setting only, where $C_s$ denotes a set of annotated source-domain sentences, and $\xi_{t}$ denotes an annotated target-domain lexicon, in which each word is associated with one POS tag. The domain adaptation model is firstly trained on $C_s$, and makes use of $\xi_{t}$ when performing domain adaptation. In practice, our method can be used in in-domain settings also.

As shown in Figure \ref{Model Architecture}, for each character $c_i$ in the input sentence, the set of all character subsequences that match words in the external lexicon $\mathbb{D}$ is denoted as $\mathbf{w}^i = \{w_{b_1, e_1}, w_{b_2, e_2}, ..., w_{b_m, e_m}\}$. Here $b_k$ and $e_k$ are the start and end index of the matched words in the sentence, where $e_k >= i$ and $b_k <= i$. Word embeddings should intuitively be used for encoding $w_{b_k, e_k}$. However, for characters forming domain specific words, there may not be readily available embeddings. POS embeddings can be used as alternative unlexicalized features of words embeddings. We introduce how to integrate POS embeddings as word information from both prediction and training.
\begin{table}[t]
	\small
	\begin{tabular}{|l|l|l|l|l|l|l|l|}
		\hline \multicolumn{2}{|c|}{Datasets}&PKU&MSR&CTB6&ZX&FR&DL\\
		\hline 
		\multirow{2}*{\begin{tabular}[c]{@{}c@{}} Training\\set\end{tabular}}&\#sent &19.1k&86.9k&23.4k & \multicolumn{3}{c|}{\multirow{3}*{PKU}}\\ 
		&\#word &1.11m&2.37m&641k  & \multicolumn{3}{c|}{\multirow{3}*{}}\\
		&\#char &1.83m&4.05m&1.06m & \multicolumn{3}{c|}{\multirow{3}*{}}\\
		\hline
		\hline
		\multirow{2}*{\begin{tabular}[c]{@{}c@{}} Testing\\set\end{tabular}} &\#sent &1.9k&4.0k& 2.8k & 0.7k & 1.3k & 1.3k\\ 
		&\#word &0.10m&0.11m&0.70m &35.2k & 35.3k& 31.5k\\
		&\#char &0.17m&0.18m&1.16m & 50.3k&64.2k & 52.2k\\
		\hline
	\end{tabular}
	\caption{Statistics of datasets}
	\label{Statistics of datasets}
\end{table}\\
\noindent\textbf{Prediction} \quad During testing, we match character subsequences in a given input sentence to a word-POS lexicon $\xi_t$. For all matching subspans, we find a vector representation by first performing a lookup action to a word embedding table, and then using the corresponding POS embedding to represent the word if no word embedding is available for the subspan.
\begin{equation}
\mathbf{x}^w_{b_k, e_k} = \begin{cases}
\mathbf{p}^w(p_i) & \text{if} \ w_{b_k, e_k} \in \xi_{t} \\
\mathbf{e}^w(w_{b_k, e_k}) &  \text{otherwise}
\end{cases} 
\end{equation}
where $\mathbf{e}^w$ and $\mathbf{p}^w$ are word and POS embedding lookup tables, respectively. \\
\noindent\textbf{Training}\quad Training is performed on a source domain corpus only. We do not fine-tune word embeddings. The key task for knowledge transfer is the learning of POS embeddings, which offer a generalized representation for words not in the embedding lexicon. To this end, we randomly replace words in the training data  with their gold-standard POS tags as follows:
\begin{equation}
P(w_{b_k, e_k}) = \text{min}(1, \sqrt{\dfrac{t}{f(w_{b_k, e_k})}})
\end{equation}
where $f(w_{b_k, e_k})$ is the frequency of $w_{b_k, e_k}$ in the training data and $t$ is a chosen threshold.

The representation of $w_{b_k, e_k}$ is:
\begin{equation}
\mathbf{x}^w_{b_k, e_k}=\begin{cases} 
\mathbf{p}^w(p_i) &  rb < P(w_{b_k, e_k})\\
\mathbf{e}^w(w_{b_k, e_k}) &   \text{otherwise}
\end{cases}
\end{equation}
where $rb$ is a random number and $p_i$ is the gold-standard POS tag of $w_{b_k, e_k}$.
\begin{figure}[t]
	\begin{flushright}
		\includegraphics[width=0.38\textwidth]{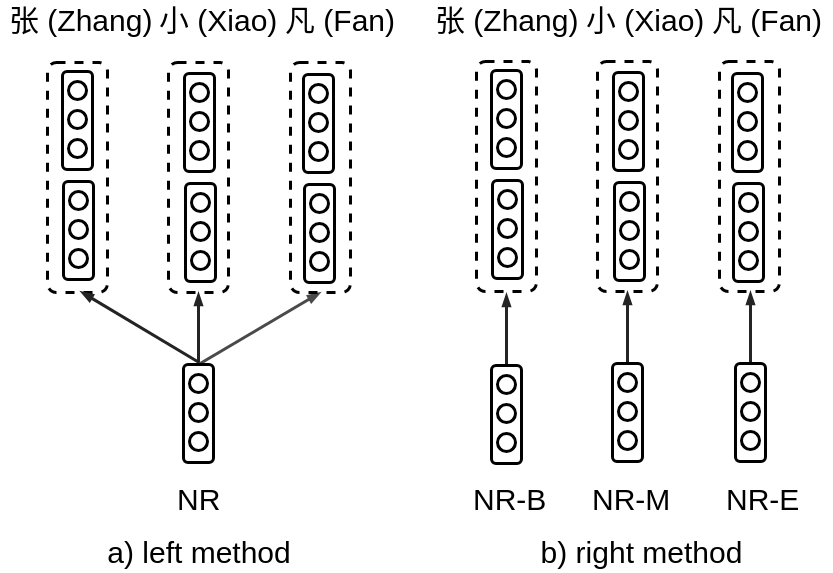}
		\caption{Two methods to learn POS embeddings. In the left method, for characters in ``张小凡(Person Name)", they attend to the same POS \textit{NR}. In the right method, different characters attend to different POS tags with positional information.}
		\label{pos}
	\end{flushright}
\end{figure}

Considering the positional information of characters in the word, the set of POS tags can be denoted in combination with segmentation labels: $P_{t\_b}=\{p_{1b}, p_{2m}, p_{1e}, p_{1s}, ..., p_{ne}\}$. The difference between $P_t$ and $P_{t\_b}$ is that for $c_j$ and matched word $w_{b_i, e_i}$, if $c_j$ is the first, middle or end character of $w_{b_i, e_i}$, the corresponding POS tag of $w_{b_i, e_i}$ is $p_{ib}$, $p_{im}$ and $p_{ie}$, respectively. Figure \ref{pos} shows the difference between $p_t$ and $p_{t\_b}$ through an example.

For each character $c_i$, we integrate dictionary word information by augmenting its embedding with a word context vector $h_i$, which is the weighted sum over $\mathbf{x}_{b_k, e_k}^w$ for all spans $(b_k, e_k)$ that contain $c_i$. In particular,
\begin{equation}
\mathbf{h}_i = \sum \alpha_{ik} \mathbf{x}_{b_k, e_k}^w
\end{equation}
where the weight for each context word is:
\begin{equation}
\begin{aligned}
\mathbf{\alpha}_{ik} & = \textrm{attention}(\mathbf{x}^c_i, \mathbf{x}^w_{b_k, e_k}) \\
& = \frac{\textrm{exp}(\textrm{score}(\mathbf{x}^c_i, \mathbf{x}^w_{b_k, e_k}))}{\sum\limits_{k=1}^{m}(\textrm{exp}(\textrm{score}(\mathbf{x}^c_i, \mathbf{x}^w_{b_k, e_k})))}
\end{aligned}
\end{equation}
Considering computation efficiency, the score function is:
\begin{equation}
\textrm{score}(\mathbf{x}^c_i, \mathbf{x}^w_{b_k, e_k}) = \mathbf{x}^c_iW \mathbf{x}^w_{b_k, e_k}
\end{equation}
where $W$ is parameters. The output of the attention layer is the concatenation of the character embedding $\mathbf{x}_i^c$ and the context vector $\mathbf{h}_i$:
\begin{equation}
\mathbf{x}_i = \mathbf{x}_i^c \oplus \mathbf{h}_i
\end{equation}

\subsection{Decoding and Training}
For decoding, the Viterbi algorithm \cite{Viterbi1967Error} is used to find the highest scored label sequence $y^*$ over a input sentence.

Given a training set with $N$ samples, the loss function is negative log-likelihood of sentence-level with $L_2$ regularization:
\begin{equation}
\textrm{Loss} = -\sum_{i=1}^{N}\textrm{log}(P(y_i|s_i)) + \frac{\lambda}{2}||\Theta||^2
\end{equation}
\begin{figure}[t]
	\centering
	\includegraphics[width=0.5\textwidth]{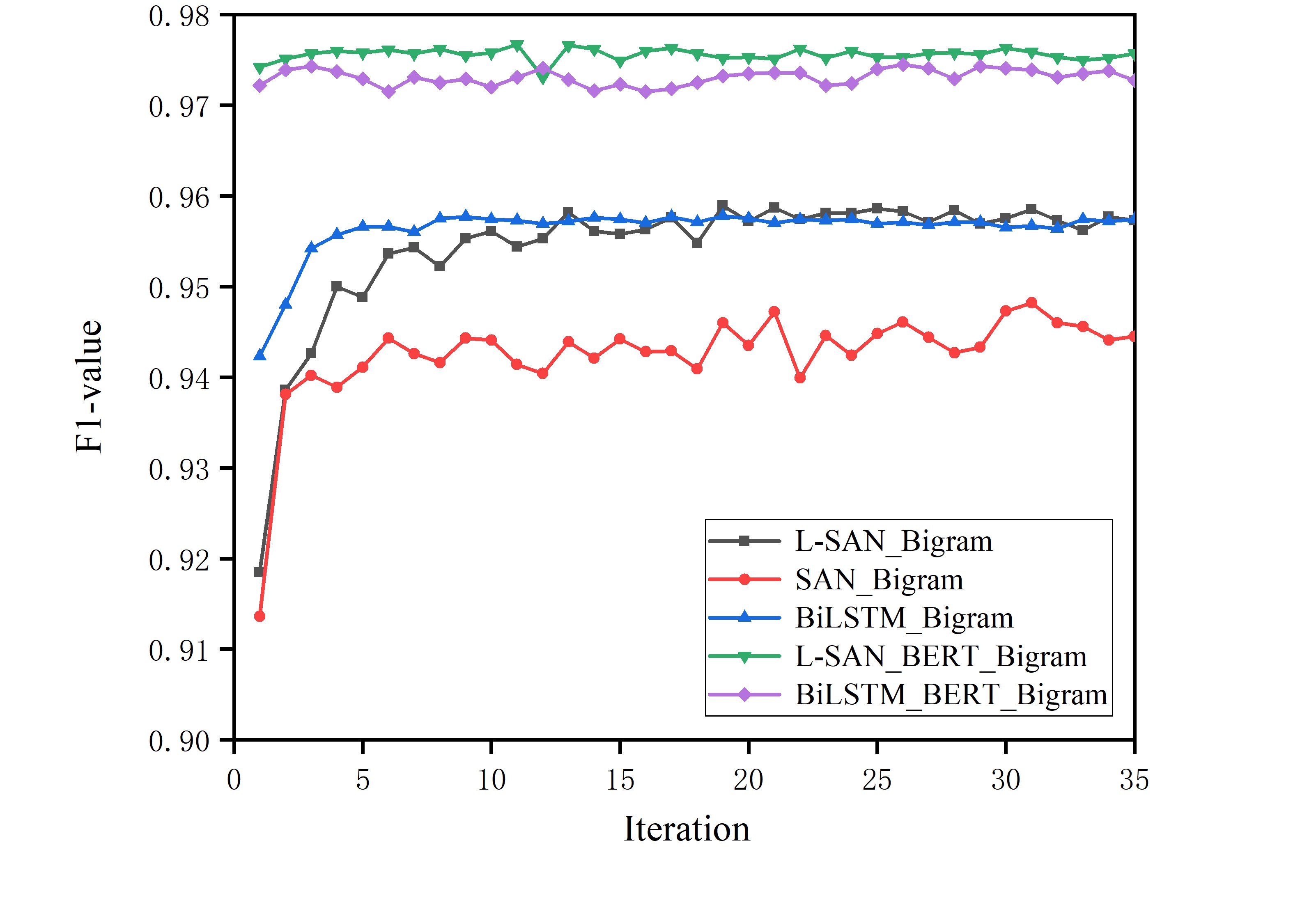}
	\caption{F1-value against training iterations}
	\label{F1-value against training iterations}
\end{figure}

\section{Experiments}
We carry out an extensive set of experiments to investigate the effectiveness of SAN-CRF and the proposed neural type-supervised domain adaptation method across different domains under different settings. F1-value is taken as our main metric. 

\subsection{Datasets}
We separately evaluate the proposed model in in-domain and cross-domain settings. For in-domain evaluation, CTB6 (Chinese Tree Bank 6.0), PKU and MSR are taken as the datasets. The train/dev/test split of CTB6 follows \citet{zhang2016transition}, while the split of PKU and MSR are taken from the SIGHAN Bakeoff 2005 \cite{emerson2005second}. For cross-domain evaluation, PKU is used as the source domain, and three Chinese novel datasets including DL (DouLuoDaLu), FR (FanRenXiuXianZhuan) and ZX (ZhuXian) \cite{qiu2015word} are used as target domains. Following \citet{zhang2014type}, we collect target-domain lexicons from Internet Encyclopedia\footnote{https://baike.baidu.com/item/诛仙/12418}\footnote{https://baike.baidu.com/item/凡人修仙传/54139}\footnote{https://baike.baidu.com/item/斗罗大陆/5313}. Table \ref{Statistics of datasets} shows the statistics of the datasets. 

\subsection{Experimental Settings}
Table \ref{Hyper-parameter values} shows the values of model hyper-parameters. For the SAN CWS model, we use the Adam \cite{kingma2014adam} optimizer with $\beta_1=0.9$, $\beta_2=0.98$, $\epsilon=10^{-9}$. Following \citet{vaswani2017attention}, we increase the learning rate linearly for the first \textit{warmup\_steps} steps, and then decrease it proportionally. The value of \textit{warmup\_steps} is set to 1000. When BERT is used for character embeddings, the learning rate is set to 5e-6. For the baseline model, we use \textbf{s}tochastic \textbf{g}radient \textbf{d}escent (SGD) follwing \citet{yang2018subword}, and the initial learning rate is set to 0.001, which gives better development results.
\begin{table}[t]
	\centering
	\small
	\begin{tabular}{|l|l||l|l|}
		\hline \textbf{Parameter}&\textbf{Value} & \textbf{Parameter} & \textbf{Value}\\
		\hline 
		Char emb size & 50 & SAN layer num & 2  \\
		Word emb size & 200  & SAN head num & 8  \\
		Bigram emb size & 50  & SAN hidden size & 512 \\
		BERT emb size & 768  & SAN Inner size & 2048 \\
		LSTM layer & 1 & SAN Relu dropout & 0.1 \\
		LSTM hidden & 200  & Attention dropout & 0.1 \\
		LSTM input dropout & 0.1  & Resiual dropout & 0.1 \\
		Batch size & 32  & Window size & 5 \\
		\hline
	\end{tabular}
	\caption{Hyper-parameter values}
	\label{Hyper-parameter values}
\end{table}
 \begin{table}[t]
	\small
	\centering
	\begin{tabular}{|c|cccccc|}
		\hline
		\textbf{\#Layer} & 1     & 2     & 3     & 4    & 5     & 6     \\ \hline
		F1                                                & 0.956 & 0.956 & 0.956 & 0.955 & 0.947 & 0.931 \\ \hline
		\hline
		\textbf{\#Head} & 2     & 4     & 6     & 8    & 12     & 16     \\ \hline
		F1                                                & 0.956 & 0.956 & 0.955 & 0.956 & 0.957 & 0.955 \\ \hline
	\end{tabular}
	\caption{Effect of numbers of heads and layers}
	\label{Number of Head and Layer}
\end{table}

\noindent \textbf{Character and Word Embedding} \quad The pre-trained word embedding size is 200, which is based on word co-occurrence and the directions of word pairs \cite{song2018directional}, and the word length is restricted to 4. we use the topmost layer output as character embedding of the pre-trained Chinese Simplified BERT model with 12 layers, 768 hidden units and 12 heads\footnote{https://github.com/huggingface/pytorch-pretrained-BERT}. Besides that, the bigram embeddings and character unigram embeddings used for attending words are the same as \citet{zhang2016transition}.

%\begin{figure}[t]
%	\centering
%	\includegraphics[width=8cm]{layernumb.jpg}
%	\caption{Number of Layer}
%	\label{Number of Layer}
%\end{figure}

%\begin{figure}[t]
%	\centering
%	\includegraphics[width=8cm]{headsnum.jpg}
%	\caption{Number of Head}
%	\label{Number of Head}
%\end{figure}

\subsection{Development Experiments}
We perform development experiments on the CTB6 development dataset to investigate the influence of hyper-parameters of self-attention network for CWS, and compare the performance of SAN, especially local self-attention, with BiLSTM. In addition, we evaluate the effect of utilizing of BERT for CWS models. 

Figure \ref{F1-value against training iterations} shows the iteration curve of F1-value against the number of training iterations with different configurations.``\_Bigram" is the model using both unigram and bigram information, and ``\_BERT" is the model replacing the word2vec character unigram representation with BERT. ``SAN" represents the original self-attention network and ``L-SAN" represents \textit{local self-attention network}. ``BiLSTM" is our baseline model, which uses a bidirectional LSTM as feature extractor.
\begin{table}[t]
	\centering
	\small
	\begin{tabular}{|l|c|c|c|}
		\hline \textbf{Models}&\textbf{CTB6}& \textbf{PKU}& \textbf{MSR}\\
		\hline 
		\citet{zhang2016transition} & 96.0 & 95.7 & 97.7\\
		\citet{cai2017fast} &- & 95.8 &97.1 \\
		\citet{yang2017neural} & 96.2 & 96.3 &97.5 \\
		\citet{zhou2017word} & 96.2 & 96.0 & 97.8 \\
		\citet{zhang2018neural} & 96.4 & 96.5 & 97.8 \\
		\citet{ma2018state} &96.7&96.1&98.1\\
		\hline
		\hline
		BiLSTM + CRF&95.2&95.1&97.2\\
		L-SAN + CRF&95.2&95.0&96.9\\
		BiLSTM + CRF + BERT&97.2&96.6&98.0\\
		L-SAN + CRF + BERT&\textbf{97.4}&\textbf{96.7}&\textbf{98.3}\\
		\hline
	\end{tabular}
	\caption{In domain results}
	\label{In domain evaluation}
\end{table}
\noindent \textbf{Width and Depth} \quad \citet{vaswani2017attention} shows that increasing the number of layers can improve the performance of English-to-German translation. We investigate the effect of number of layers on CWS, by increasing the number of layers from 1 to 6 while fixing the number of heads to 8. The results are listed in Table \ref{Number of Head and Layer}. The model achieves the best F1-value 0.956 within 3 layers, after which the performance decreases with the increasing of layers. The F1-value decreases to 0.931 when the number of layers is set to 6. We fix the number of layers to 2 for the remaining experiments.

We vary the number of heads in multi-head self-attention to investigate its effect on CWS performance. The number of layers and dimension of head is fix to 2 and 64, respectively. As shown in Table \ref{Number of Head and Layer}, with increasing number of heads from 2 to 16, the performance does not vary too much. We fix the number of heads to 8 for the remaining experiments.

\noindent \textbf{Effect of Local Attention}\quad As Figure \ref{F1-value against training iterations} shows, the performance of ``L-SAN\_Bigram" gives much better results compared to ``SAN\_Bigram", which suggests that long-term dependencies can bring more noise than useful information. The proposed local self-attention network model can achieve the competitive results comparing with the baseline BiLSTM model. 

\noindent \textbf{Effect of BERT}\quad By replacing word2vec character embeddings with BERT, both BiLSTM and L-SAN models can reach the best F1-value within several epochs, with a significant improvement, which proves that context-dependent word representation can benefit CWS task.
\begin{table}[t]
	\centering
	\small
	\begin{tabular}{|l|c|c|c|}
		\hline
		\textbf{Model}&\textbf{ZX}&\textbf{FR}&\textbf{DL}\\
		\hline 
		\citet{liu2012unsupervised}&87.2&87.5&91.4\\
		
		\citet{qiu2015word}&87.4&86.7&91.9\\
		
		%\cite{huang2017addressing}& - & - & - &93.99&95.81&94.33&92.26\\
		%\cite{zhao2018neural}& - & - & - &95.32&95.84&93.23&93.73\\
		%\cite{zhang2018neural}& - & - & - &94.70&96.06&94.76&94.18\\
		\citet{ye2019improving}&89.6&89.6&93.5\\
		\hline
		\hline
		L-SAN + CRF + BERT&90.5&91.1&93.0\\
		L-SAN + CRF + BERT + \textit{t}&91.8&92.3&94.3\\
		L-SAN + CRF + BERT + \textit{t}\_\textit{b}&\textbf{93.1}&\textbf{93.0}&\textbf{95.1}\\
		\hline
	\end{tabular}
	\caption{Cross domain results}
	\label{Cross domain evaluation}
\end{table}
\subsection{Final Results}
\noindent \textbf{In-Domain Results}\quad We evaluate our model on three news datasets, including CTB, PKU and MSR. The main results and the recent state-of-the-art models are listed in Table \ref{In domain evaluation}. Compared with the baseline ``BiLSTM+CRF" model, the proposed ``L-SAN+CRF" model can achieve similar results, which proves that self-attention network can be a competitive feature extractor for CWS besides recurrent neural network. When replacing word2vec character embedding with BERT, the ``BiLSTM+CRF" model gives 41.3\%, 30.6\% and 31.0\% error reduction on CTB6/PKU/MSR, respectively,  and the ``L-SAN+CRF" model has 41.3\%, 32.7\% and 41.3\% error reductions on three in-domain datasets, respectively. Finally, ``L-SAN+CRF" slightly outperforms ``BiLSTM+CRF" when using BERT as unigram character representation.

\noindent \textbf{Cross-Domain Results}\quad We evaluate our model on the three cross-domain datasets, including ZX, FR and DL. The main results and three state-of-the-art models are listed in Table \ref{Cross domain evaluation}. ``\textit{t}" means neural type-supervised method is used to learn POS embedding and domain-specific words is generalized to corresponding tag. In ``\textit{t}\_\textit{b}", we learn different POS embeddings for different positions in a word. 

As shown in Table \ref*{Cross domain evaluation}, the F1-value of ``L-SAN+CRF+BERT" has an average 0.7 improvement compared with the state-of-the-art results \cite{ye2019improving} in ZX and FR without using \citet{ye2019improving}'s domain adaptation techniques. This may be because ZX, FR and DL are all Chinese novels which contain a large number of noun entities and their wring styles are different from news domain. The result shows that BERT has rarely less effect on cross-domain CWS compared with strong domain adaptation methods. The ``L-SAN+CRF+BERT+\textit{t}" model has 21.15\%, 25.96\% and 1.54\% error reduction on ZX/FR/DL datasets, respectively, which shows that the proposed neural type-supervised method can handle out of vocabulary words more effective. For characters within a word, instead of sharing the same POS embedding of the word, we further distinguish POS embeddings of characters according to their position in a word. The ``L-SAN+CRF+BERT+\textit{t}\_\textit{b}" gives 33.65\%, 32.69\% and 24.62\% on three datasets, respectively. We believe that this is due to more supervision information.
\begin{figure}[t]
	\centering
	\includegraphics[width=8cm]{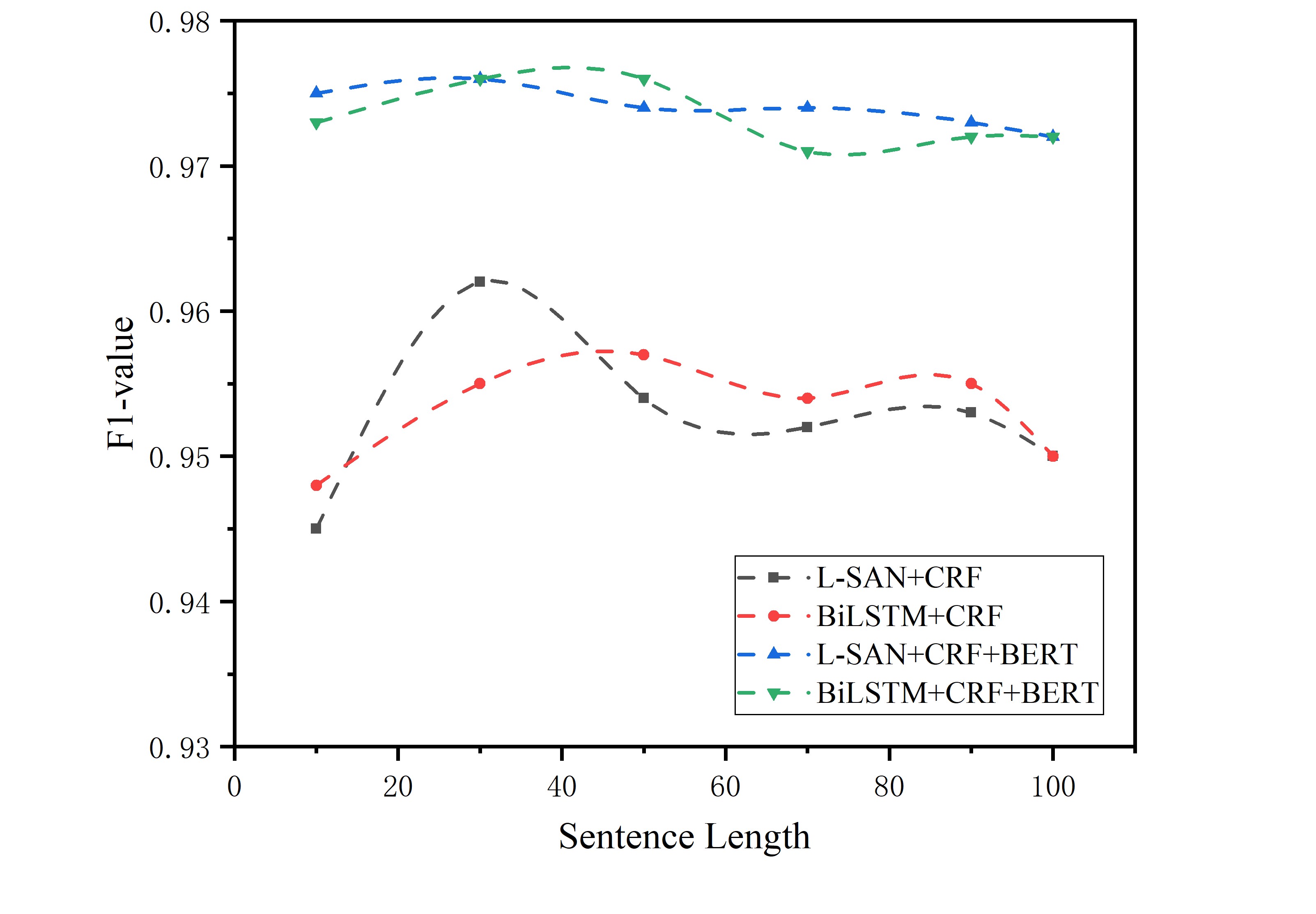}
	\caption{F1-value against the sentence length}
	\label{Sentence Length}
\end{figure}
\subsection{Analysis}
\noindent \textbf{Sentence Length}\quad We compare the baseline model and local self-attention network model, as well as the two models with BERT input representation on different sentence lengths. Figure \ref{Sentence Length} shows the F1-value on CTB6 test dataset. The two models without using BERT show a similar performance-length curve, which reaches a peak at around 30-character sentences and decreases when sentence length over 90. One possible reason is that very short sentences are rare while long sentences are semantically more challenging. However, the two models using BERT both show more stable performance-length curves, which shows that contextualized BERT representation can stabilize performance against sentence length.
\begin{table}[t]
	\small	
	\begin{tabular}{|l|c|c|c|c|}
		\hline
		\textbf{Word} & \textbf{Count}  & \textbf{M1} & \textbf{M2} & \textbf{M3} \\ \hline
		唐三(person name)  & 273 & 0.98            & 0.99             & 1.00                 \\
		韩立(person name)  & 185 & 0.07            & 0.67             & 1.00                 \\
		戴沐白(person name)   & 159 & 0.01           & 0.31              & 1.00                 \\
		小舞(person name)  & 153 & 0.90           & 0.98             & 1.00                 \\
		张小凡(person name) & 142 & 0.00           & 0.06             & 1.00                 \\
		玄骨(person name)  & 114 & 0.96            & 0.97             & 0.98                 \\
		魂狮(proper name)  & 90 & 1.00            & 1.00             & 1.00                 \\
		宁荣荣(person name)  & 86 & 0.01           & 0.57              & 1.00                 \\
		朱竹清(person name)  & 81 & 0.03           & 0.76             & 1.00                 \\
		魂环(proper name) & 72 & 1.00           & 1.00             & 1.00                 \\
		魂力(proper name)  & 71 & 0.97            & 0.99             & 1.00                 \\
		魂兽(proper name)  & 53 & 1.00            & 1.00             & 1.00                 \\
		斗魂(proper name)   & 51 & 0.71           & 0.73              & 0.76                 \\
		叶知秋(person name)  & 51 & 0.00           & 0.00             & 0.00                 \\
		乌丑(person name) & 45 & 0.97           & 1.00             & 1.00                 \\ \hline
		average precision & 108 & 0.55          & 0.73             &0.96                 \\ \hline
	\end{tabular}
	\caption{Segmentation precision of noun entities with the highest frequency.}
	\label{Noun Entity}
\end{table}

\noindent \textbf{Noun Entity Segmentation}\quad Noun entities raise a key problem for cross-domain CWS. Table \ref{Noun Entity} shows the three models segmentation results on 15 noun entities with the highest frequency of three datasets. M1 and M2 represent ``L-SAN+CRF+BERT" and ``L-SAN+CRF+BERT+\textit{t}", respectively, while M3 represents ``L-SAN+CRF+BERT+\textit{t\_b}". As the table shows, the average precision of MI is 0.55. By using neural type-supervised domain adaptation method, the average precision of M2 has a improvement of 0.18 in absolute value.

Some person names are incorrectly segmented by M2, such as ``戴沐白(person name)" and ``张小凡(person name)". When incorporating the positional information of character in the word, the average segmentation precision improves further and most noun entities can be correctly segmented, except the word ``叶知秋(person name)". The reason is that the domain lexicon does not contain ``叶知秋". This shows that our method makes effective use of domain lexicons.\\
\noindent \textbf{Case Study}\quad We use two examples of neural type-supervised domain adaptation for illustrated discussion. In example 1, ``L-SAN+CRF+BERT" fails to handle the domain entity noun ``韩立(person name)" while the two neural type-supervised domain adaptation method segment it correctly. For example 3, only ``L-SAN+CRF+BERT+\textit{t\_b}" segments it correctly. One possible reason is that it maybe difficult to distinguish between ``戴沐白 (person name)", which is a domain specific entity noun and ``白虎 (white tiger)", which is a common noun.

\section{Related Work}
%几种基于神经网络的中文分词方法: character/word/pretrain
\noindent\textbf{Chinese Word Segmentation}\quad \citet{chen2015long, chen2015gated} extract features based on character representation by using LSTM or GRU. \citet{zhang2016transition} propose a transition-based neural model which can utilize the word-level features. \citet{zhou2017word} trains character embedding with word-based context information on auto-segmented data. \citet{yang2017neural} exploit the effectiveness of rich external resources through multi-task learning. For cross-domain CWS, \citet{zhang2014type} propose a type-supervised domain adaptation approach for joint CWS and POS-tagging, which shows a competitive result compared to token-supervised methods. \citet{qiu2015word} investigate CWS for Chinese novels, proposing a method which can automatically mine noun entities for novels using a double-propagation algorithm. \citet{zhang2018neural} investigate how to integrate external dictionary into CWS models. Similar to \citet{zhang2014type} and \citet{zhang2018neural}, our work uses domain lexicon. The difference is we utilize POS embeddings through an end-to-end neural method.
% cross domain cws

\noindent\textbf{Self-Attention Network}\quad Self-attention network \cite{vaswani2017attention} was first proposed for machine translation. \citet{tan2018deep} and \citet{strubell2018linguistically} use SAN for the task of semantic role labeling, which can directly capture the relationship between two arbitrary tokens in the sequence. \citet{strubell2018linguistically} incorporate linguistic information through multi-task learning, including dependency parsing, part-of-speech and predicate detection. \citet{shen2018disan} propose multi-dimensional attention as well as directional information, achieving the state-of-the-art results on natural language inference and sentiment analysis tasks. \citet{kitaev2018constituency} show that a novel encoder based on self-attention can lead to state-of-the-art results for the constituency parsing task. Along with this strand of work, we study the influence of global and local attention for CWS and build a SAN-CRF word segmentor, which gives competitive results compared with BiLSTMs.
\begin{table}[t]
	\tiny
	\begin{tabular}{|l|l|}
		\hline
		\multicolumn{2}{|l|}{\begin{tabular}[c]{@{}l@{}}\#Example 1: 韩立也在光罩边缘处止住了下落的身影\\ \quad \quad \quad \quad \quad \ Han Li also stopped the falling figure at the edge of the mask\end{tabular}}    \\ \hline
		Gold Segmentation                            & \begin{tabular}[c]{@{}l@{}}韩立/也/在/光罩/边缘/处/止住/了/下落/的/身形\\ Han Li/also/at/the mask/the edge/of/stopped/x/the falling/figure\end{tabular}  \\ \hline
		L-SAN+CRF+BERT                               & \begin{tabular}[c]{@{}l@{}}韩/立/也/在/光罩/边缘/处/止住/了/下落/的/身形\\ Han/Li/also/at/the mask/the edge/of/stopped/x/the falling/figure\end{tabular} \\ \hline
		L-SAN+CRF+BERT+\textit{t}   & \begin{tabular}[c]{@{}l@{}}韩立/也/在/光罩/边缘/处/止住/了/下落/的/身形\\ Han Li/also/at/the mask/the edge/of/stopped/x/the falling/figure\end{tabular}  \\ \hline
		L-SAN+CRF+BERT+\textit{t\_b} & \begin{tabular}[c]{@{}l@{}}韩立/也/在/光罩/边缘/处/止住/了/下落/的/身形\\ Han Li/also/at/the mask/the edge/of/stopped/x/the falling/figure\end{tabular}  \\ \hline
		\multicolumn{2}{|l|}{\begin{tabular}[c]{@{}l@{}}\#Example 2: 戴沐白虎掌上利刃弹开\\ \quad \quad \quad \quad \quad \ Dai Mubai pops up the blade on the palm\end{tabular}}                                  \\ \hline
		Gold Segmentation                            & \begin{tabular}[c]{@{}l@{}}戴沐白/虎掌/上/利刃/弹开\\ Dai Mubai/palm/on/blade/pops up\end{tabular}                                                \\ \hline
		L-SAN+CRF+BERT                               & \begin{tabular}[c]{@{}l@{}}戴/沐/白虎/掌/上/利刃/弹开\\ Dai/Mu/white tiger/palm/on/blade/pops up\end{tabular}                                     \\ \hline
		L-SAN+CRF+BERT+\textit{t}   & \begin{tabular}[c]{@{}l@{}}戴沐/白虎/掌上/利刃/弹开\\ Dai Mu/white tiger/palm/on/blade/pops up\end{tabular}                                       \\ \hline
		L-SAN+CRF+BERT+\textit{t\_b} & \begin{tabular}[c]{@{}l@{}}戴沐白/虎掌/上/利刃/弹开\\ Dai Mubai/palm/on/blade/pops up\end{tabular}                                                \\ \hline
	\end{tabular}
	\caption{Examples. x represents ungrammatical word.}
\end{table}
%上下文相关词向量的介绍，GPT，ELMo, BERT

\noindent\textbf{Contextualized word representation}\quad Context-dependent word representations pre-trained from large-scale corpora have received much recent attention. ELMo \cite{peters2018deep} is based on recurrent neural networks language models. OpenAI GPT \cite{radford2018improving} builds a left-to-right language model with a multi-layer multi-head self-attention networks, which can handle long-term dependencies better compared to recurrent networks. Different from OpenAI GPT, BERT \cite{devlin2018bert} uses a deep bidirectional Transformer pre-trained on Masked LM. Our work investigates the effect of contextualized character representation on both in-domain and cross-domain CWS under a unified SAN framework.

\section{Conclusion}
We investigated self-attention network for Chinese word segmentation, demonstrating that it can achieve comparable results with recurrent network methods. We found that local attention gives better results compared to standard SAN. Under SAN, we also investigate the influence of rich character and word features, including BERT character embeddings and a neural attention method to integrate word information into character based CWS. Extensive in-domain and cross-domain experiments show that the proposed SAN method archives state-of-the-art performance on both in-domain and cross-domain Chinese word segmentation datasets.

%\bibliography{acl2018}
\bibliography{acl2018}
\bibliographystyle{acl_natbib}
\end{CJK*}
\end{document}